\documentclass[final,3pt,times,twocolumn]{elsarticle}
\usepackage{amsmath,amssymb,amsfonts}
\usepackage{hyperref}
\usepackage{url}
\usepackage{nicefrac}
\usepackage{subcaption}
\usepackage{booktabs}
\usepackage{algorithmic}
\usepackage{graphicx}
\usepackage{textcomp}
\usepackage{xcolor}
\usepackage{soul}

\usepackage{enumitem}
\usepackage{caption}
\usepackage{xurl}
\usepackage{longtable}
\usepackage{multirow}
\usepackage{xcolor}

\journal{Artificial Intelligence in Medicine}

\begin{document}

\begin{frontmatter}

\title{Multi-domain Clinical Natural Language Processing with MedCAT: the Medical Concept Annotation Toolkit}
\author[inst1]{Zeljko Kraljevic\fnref{contrib}}
\author[inst1,inst6]{Thomas Searle\fnref{contrib}}
\author[inst3]{Anthony Shek}
\author[inst2,inst4,inst8]{Lukasz Roguski}
\author[inst2,inst4,inst8]{Kawsar Noor}
\author[inst1,inst2]{Daniel Bean}
\author[inst1,inst6]{Aurelie Mascio}
\author[inst4,inst8]{Leilei Zhu}
\author[inst1,inst4,inst6]{Amos A. Folarin}
\author[inst1,inst2,inst6]{Angus Roberts}
\author[inst1,inst6]{Rebecca Bendayan}
\author[inst3]{Mark P. Richardson}
\author[inst5,inst6]{Robert Stewart}
\author[inst2,inst4,inst8]{Anoop D. Shah}
\author[inst4,inst8]{Wai Keong Wong}
\author[inst1]{Zina Ibrahim}
\author[inst3,inst7]{James T. Teo}
\author[inst1,inst2,inst4,inst6]{Richard JB. Dobson\corref{cor}}
\affiliation[inst1]{organisation={Department of Biostatistics and Health Informatics Institute of Psychiatry, Psychology and Neuroscience, King’s College London,},city={London},country={U.K.}}
\affiliation[inst2]{organisation={Health Data Research UK London, University College London,},city={London},country={U.K.}}
\affiliation[inst3]{organisation={Department of Clinical Neuroscience, Institute of Psychiatry, Psychology and Neuroscience, King’s College London,},city={London},country={U.K.}}
\affiliation[inst4]{organisation={Institute of Health Informatics, University College London,}city{London,},country={U.K.}}
\affiliation[inst5]{organisation={Department of Psychological Medicine, Institute of Psychiatry, Psychology and Neuroscience, King’s College London,},city={London,},country={U.K.}}
\affiliation[inst6]{organisation={NIHR Biomedical Research Centre at South London and Maudsley NHS Foundation Trust and King’s College London,},city={London},country={U.K.}}
\affiliation[inst7]{organisation={Department of Neurology, King’s College Hospital NHS Foundation Trust,},city={London,},country={U.K.}}
\affiliation[inst8]{organisation={NIHR BRC Clinical Research Informatics Unit, University College London Hospitals, NHS Foundation Trust}, city={London}, country={U.K.}}
\fntext[contrib]{Authors contributed equally}
\cortext[cor]{Corresponding author: richard.j.dobson@kcl.ac.uk}

\begin{abstract}
Electronic health records (EHR) contain large volumes of unstructured text, requiring the application of Information Extraction (IE) technologies to enable clinical analysis. We present the open source Medical Concept Annotation Toolkit (MedCAT) that provides: a) a novel self-supervised machine learning algorithm for extracting concepts using any concept vocabulary including UMLS/SNOMED-CT; b) a feature-rich annotation interface for customising and training IE models; and c) integrations to the broader CogStack ecosystem for vendor-agnostic health system deployment. We show improved performance in extracting UMLS concepts from open datasets (F1:0.448-0.738 vs 0.429-0.650). Further real-world validation demonstrates SNOMED-CT extraction at 3 large London hospitals with self-supervised training over $\sim$8.8B words from $\sim$17M clinical records and further fine-tuning with ~6K clinician annotated examples. We show strong transferability (F1\textgreater0.94) between hospitals, datasets and concept types indicating cross-domain EHR-agnostic utility for accelerated clinical and research use cases.
\end{abstract}


\begin{keyword}
Electronic Health Record Information Extraction\sep Clinical Natural Language Processing \sep Clinical Concept Embeddings \sep Clinical Ontology Embeddings 
\end{keyword}

\end{frontmatter}

\section{Introduction}\label{sec:introduction}
Electronic Health Records (EHR) are large repositories of clinical and operational data that have a variety of use cases from population health, clinical decision support, risk factor stratification and clinical research. However, health record systems store large portions of clinical information in unstructured format or proprietary structured formats, resulting in data that is hard to manipulate, extract and analyse. There is a need for a platform to accurately extract information from freeform health text in a scalable manner that is agnostic to underlying health informatics architectures.

We present the Medical Concept Annotation Toolkit (MedCAT): an open-source Named Entity Recognition + Linking (NER+L) and contextualization library, an annotation tool and online learning training interface, and integration service for broader CogStack\cite{Jackson2018-km} ecosystem integration for easy deployment into health systems. The MedCAT library can learn to extract concepts (e.g. disease, symptoms, medications) from free-text and link them to any biomedical ontology such as SNOMED-CT\cite{Stearns2001-yt} and UMLS\cite{Bodenreider2004-ci}. MedCATtrainer\cite{Searle2019-rq}, the annotation tool, enables clinicians to inspect, improve and customize the extracted concepts via a web interface built for training MedCAT information extraction pipelines. This work outlines the technical contributions of MedCAT and compares the effectiveness of these technologies with existing biomedical NER+L tools. We further present real clinical usage of our work in the analysis of multiple EHRs across various NHS hospital sites including running the system over ~20 years of collected data pre-dating even the usage of modern EHRs at one site. MedCAT has been deployed and contributed to clinical research findings in multiple NHS trusts throughout England\cite{Bean2020-vq,Carr2020-tj}.

\subsection{Problem Definition}
\begin{figure*}
    \centering
    \includegraphics[scale=0.25]{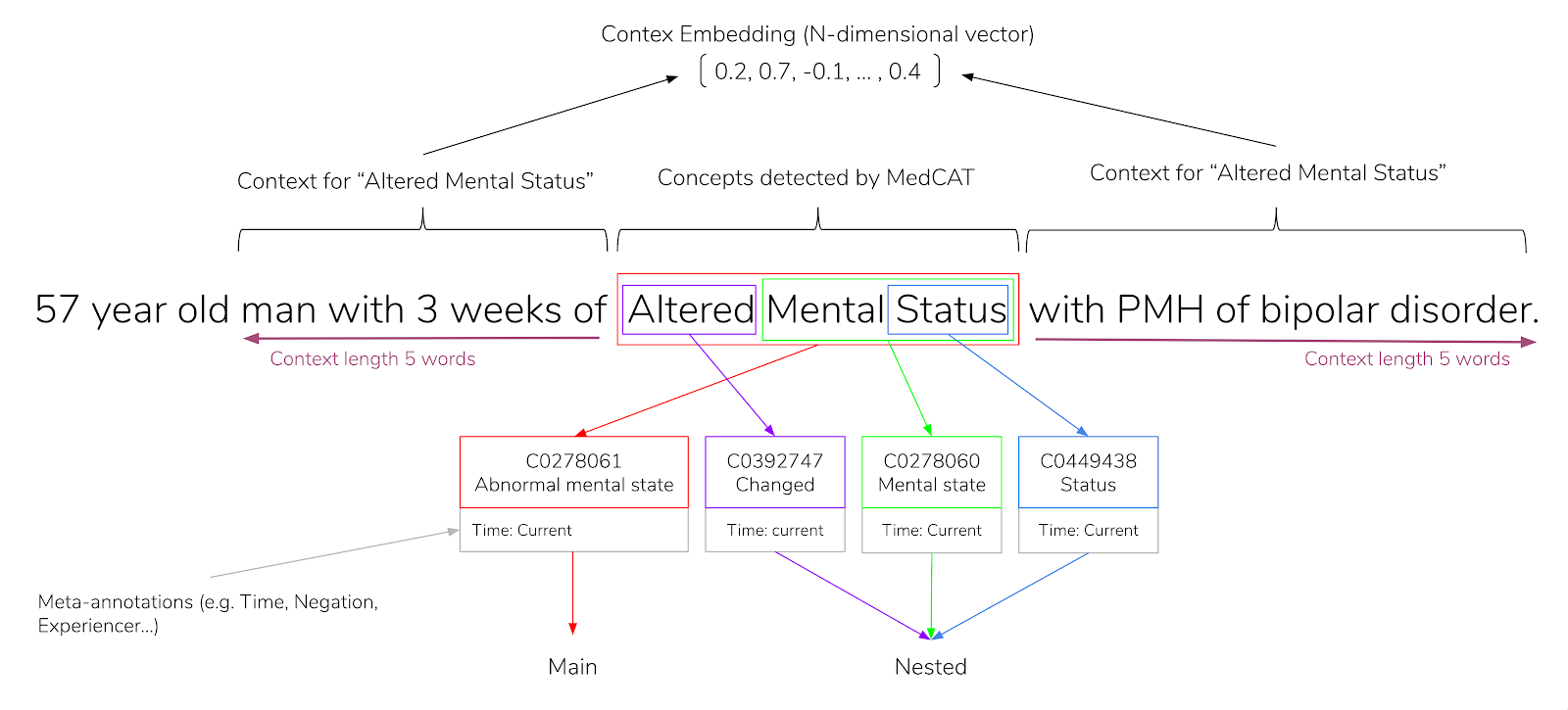}
    \caption{A fictitious example of biomedical NER+L with nested entities and further ‘meta-annotations’; a further classification or `context' applied to an already extracted concept e.g. ‘time current’ indicates extracted concepts are mentioned in a temporally present context. This context may also be referred to as an attribute of a recognised entity. Each one of the detected boxes (nested) has multiple candidates in the Unified Medical Language System (UMLS). The goal is to detect the entity and annotate it with the most appropriate concept ID, e.g. for the span Status, we have at least three candidates in UMLS, namely C0449438, C1444752, C1546481.}
    \label{fig:example_ner_l}
\end{figure*}

Recently NER models based on Deep Learning (DL), notably Transformers\cite{Vaswani2017-db} and Long-Short Term Memory Networks\cite{Hochreiter1997-uw} have achieved considerable improvements in accuracy\cite{Howard2018-ec}. However, both approaches require explicit supervised training. In the case of biomedical concept extraction, there is little publicly available labelled data due to the personal and sensitive nature of the text. Building such a corpus can be onerous and expensive due to the need for direct EHR access and domain expert annotators. In addition, medical vocabularies can contain millions of different named entities with overlaps (see Fig. \ref{fig:example_ner_l}). Extracted entities will also often require further classification to ensure they are contextually relevant; for example extracted concepts may need to be ignored if they occurred in the past or are negated. We denote this further classification as meta-annotation or a `contextualisation' of a recognised entity. Overall, using data-intensive methods such as DL can be extremely challenging in real clinical settings. 

This work is positioned to improve on current tools such as the Open Biomedical Annotator (OBA) service\cite{Jonquet2009-qc} that have been used in tools such as DeepPatient\cite{Miotto2016-ij} and ConvAE\cite{Landi2020-cd} to structure and infer clinically meaningful outputs from EHRs. MedCAT allows for continual improvement of annotated concepts through a novel self-supervised machine learning algorithm, customisation of concept vocabularies, and downstream contextualisation of extracted concepts. All of which are either partially or not addressed by current tools.

\subsection{NER+L in a Biomedical Context}
Due to the limited availability of training data in biomedical NER+L, existing tools often employ a dictionary-based approach. This involves the usage of a vocabulary of all possible terms of interest and the associated linked concept as specified in the clinical database e.g. UMLS or SNOMED-CT.
This approach allows the detection of concepts without providing manual annotations. However, it poses several challenges that occur frequently in EHR text. These include: spelling mistakes, form variability (e.g. kidney failure vs failure of kidneys), recognition and disambiguation (e.g. does ‘hr’ refer to the concept for ‘hour’ or ‘heart rate’ or neither).

\subsection{Existing Biomedical NER+L Tools}\label{sec:existing_tools}
We compare prior NER+L tools for biomedical documents that are capable of handling extremely large concept databases (completely and not a small subset).
MetaMap\cite{Aronson2010-rw} was developed to map biomedical text to the UMLS Metathesaurus. MetaMap cannot handle spelling mistakes and has limited capabilities to handle ambiguous concepts. It offers an opaque additional ‘Word-Sense-Disambiguation’ system that attempts to disambiguate candidate concepts that consequently slows extraction.
Bio-YODIE\cite{Gorrell2018-ub} improves upon the speed of extraction compared to MetaMap and includes improved disambiguation capabilities, but requires an annotated corpus or supervised training. SemEHR\cite{Wu2018-sm} builds upon Bio-YODIE to somewhat address these shortcomings by applying manual rules to the output of Bio-YODIE to improve the results. Manual rules can be labour-intensive, brittle and time-consuming, but they can produce good results\cite{Gorinski2019-cr}.
cTAKES\cite{Savova2010-df}, builds on existing open-source technologies—the Unstructured Information Management Architecture\cite{Ferrucci2004-ob} framework and OpenNLP\cite{Morton2005-rf} the natural language processing toolkit. The core cTAKES library does not handle any of the previously mentioned challenges without additional plugins.
ScispaCy\cite{Neumann2019-wi} is a practical biomedical/scientific text processing tool, which heavily leverages the spaCy\footnote{https://github.com/explosion/spaCy} library. In contrast to other tools mentioned, ScispaCy is primarily a supervised model for NER with limited linking capabilities. CLAMP\cite{soysal} is a comprehensive clinical NLP software that enables recognition and automatic encoding of clinical information in narrative patient reports. Similar to ScispaCy it is a supervised approach and not directly comparable to other tools mentioned here.
MetaMap, BioYODIE, SemEHR, cTakes and ScispaCy only support extraction of UMLS concepts. BioPortal\cite{Whetzel2011-bm} offers a web hosted annotation API for 880 distinct ontologies. This is important for use cases that are not well supported by only the UMLS concept vocabulary\cite{Keselman2008-ck} or are better suited to alternative terminologies\cite{Wang2018-hr}. However, transmitting sensitive hospital data to an externally hosted annotation web API may be prohibited under data protection legislation\cite{noauthor_undated-ls}. The BioPortal annotator is a ‘fixed’ algorithm so does not allow customisation or improvements through machine learning or support of non-english language corpora\cite{Hellrich2014-mi}. 

CLAMP, and in a limited capacity cTakes and SemEHR, support further contextualisation of extracted concepts. MetaMap, BioYODIE and scispaCy treat this as a downstream task although it is often required before extracted concepts can be used in clinical research. MedCAT addresses these shortcomings of prior tools allowing for flexibly clinician driven definition of concept contextualisation, supporting modern information extraction requirements for biomedical text.

\section{Methods}
\begin{figure*}
    \centering
    \includegraphics[scale=0.35]{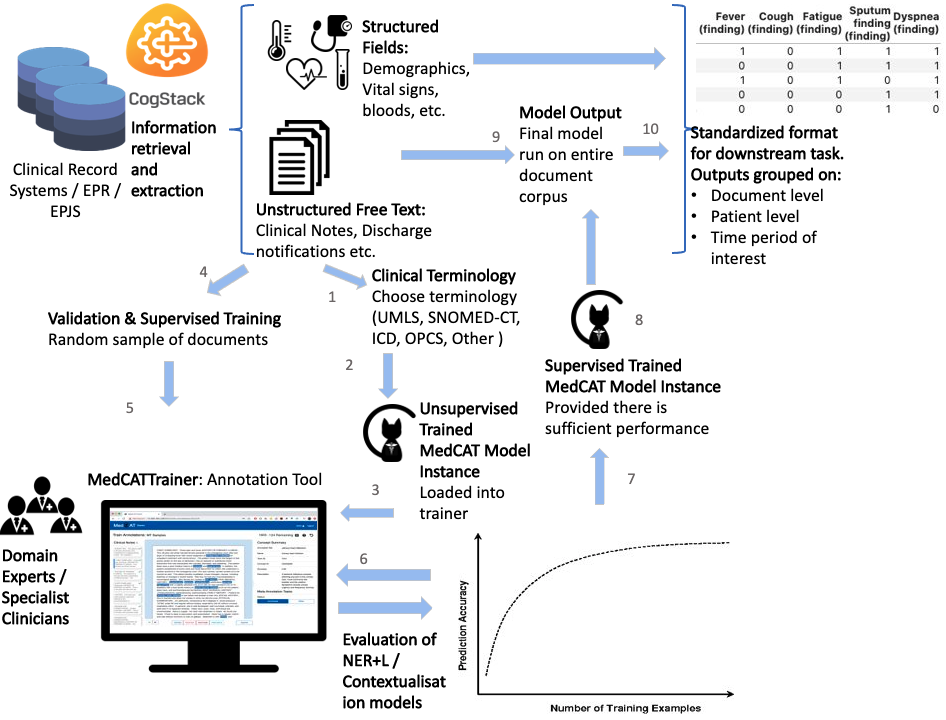}
    \caption{An example MedCAT workflow using the MedCAT core library and MedCATtrainer technologies to support clinical research.}
    \label{fig:medcat_within_cogstack_ecosystem}
\end{figure*}

MedCAT presents a set of decoupled technologies for developing IE pipelines for varied health informatics use cases. Fig. \ref{fig:medcat_within_cogstack_ecosystem} shows a typical MedCAT workflow within a wider typical CogStack deployment. CogStack queries selectively extract relevant documents from the EHR including the structured and unstructured (freetext) notes. With MedCAT we firstly agree with clinical partners the relevant terms within a clinical terminology(1) and train MedCAT self-supervised(2). We load the model into the MedCATtrainer annotation tool(3) alongside a random sample of the extracted EHR documents(4). Clinical domain experts validate and improve the model using supervised online learning(5). Metrics demonstrate the quality of a fine-tuned MedCAT model(6) and once desired performance is reached the fine-tuned model is exported(7) and run upon the wider free-text EHR dataset(8,9), facilitating downstream clinical research through the newly structured data(10).

This section presents the MedCAT platform technologies, its method for learning to extract and contextualise biomedical concepts through self-supervised and supervised learning. Integrations with the broader CogStack ecosystem are presented alongside source code\footnote{\url{https://cogstack.atlassian.net/wiki/spaces/COGDOC/pages/733380653/Natural+Language+Processing}}.  Finally, we present our experimental methodology for assessing MedCAT in real clinical scenarios.

\subsection{The MedCAT Core Library}\label{medcat_core_lib}
We now outline the technical details of the NER+L algorithm, the self-supervised and supervised training procedures and methods for flexibly contextualising linked entities. 

\subsubsection{Vocabulary and Concept Database}
MedCAT NER+L relies on two core components: 
\begin{itemize}[leftmargin=*]
    \item \textbf{Vocabulary (VCB)}: the list of all possible words that can appear in the documents to be annotated. It is primarily used for the spell checking features of the algorithm. We have compiled our own VCB by scraping Wikipedia and enriching it with words from UMLS. Only the Wikipedia VCB is made public, but the full VCB can be built with scripts provided in the MedCAT repository (https://github.com/CogStack/MedCAT). The scripts require access to the UMLS Metathesaurus (https://www.nlm.nih.gov/research/umls).
    \item \textbf{Concept Database (CDB)}: a table representing a biomedical concept dictionary (e.g. UMLS, SNOMED-CT). Each new concept added to the CDB is represented by an ID and Name. A concept ID can be referred to through multiple names with identical conceptual meanings such as heart failure, myocardial failure, weak heart and cardiac failure.
\end{itemize} 

\subsubsection{The NER+L Algorithm}
With a prepared CDB and VCB, we perform a first pass NER+L pipeline then run a trainable disambiguation algorithm. The initial NER+L pipeline starts with cleaning and spell-checking the input text. We employ a fast and lightweight spell checker (http://www.norvig.com/spell-correct.html) that uses word frequency and edit distance between misspelled and correct words to fix mistakes. We use the following rules: 

\begin{itemize}[leftmargin=*]
    \item A word is spelled against the VCB, but corrected only against the CDB.
    \item The spelling is never corrected in the case of abbreviations.
    \item An increase in the word length corresponds to an increase in character correction allowance.
\end{itemize}

Next, the document is tokenized and lemmatized to ensure a broader coverage of all the different forms of a concept name. We used SciSpaCy\cite{Neumann2019-wi}, a tool tuned for these tasks in the biomedical domain. Finally, to detect entity candidates we use a dictionary-based approach with a moving expanding window:

\begin{enumerate}[leftmargin=*]
    \item Given a document ${d_1}$
    \item Set window\_length = 1 and word\_position = 0 
    \item There are three possible cases: 
    \begin{enumerate}[leftmargin=*]
        \item The text in the current window is a concept in our CDB (the concept dictionary), mark it and go to 4. Note that MedCAT can ignore token order, but only for up-to two tokens (stopwords are not counted in the two token limit).
        \item The text is a substring of a longer concept name, if so go to 4.
        \item Otherwise reset window\_length to 1, increase word\_position by 1 and repeat step 3 
    \end{enumerate}
    \item Expand the window size by 1 and repeat 3.
\end{enumerate}

Steps 3 and 4 help us solve the problem of overlapping entities shown in Fig. \ref{fig:example_ner_l}. 

\subsection{Self-Supervised Training Procedure} 
For concept recognition and disambiguation, we use context similarity. Initially, we find and annotate mentions of concepts that are unambiguous, (e.g. step 3. a. in the previous expanding window algorithm) then we learn the context of marked text spans. For new documents, when a concept candidate is detected and is ambiguous its context is compared to the currently learned one, if the similarity is above a threshold the candidate is annotated and linked. The similarity between the context embeddings also serves as a confidence score of the annotation and can be later used for filtering and further analysis. 
The self-supervised training procedure is defined as follows: 
\begin{enumerate}[leftmargin=*]
    \item Given a corpus of biomedical documents and a CDB.
    \item For each concept in the CDB ignore all names that are not unique (ambiguous) or that are known abbreviations.
    \item Iterate over the documents and annotate all of the concepts using the approach described earlier. The filtering applied in the previous steps guarantee the entity can be annotated.
    \item For each annotated entity calculate the context embedding $V_{cntx}$. 
    \item Update the concept embedding $V_{concept}$ with the context embedding $V_{cntx}$.
\end{enumerate}

The self-supervised training relies upon one of the names assigned to each concept to be unique in the CDB. The unique name is a reference point for training to learn concept context, so when an ambiguous name appears (a name that is used for more than one concept in the CDB) it can be disambiguated. For example, the UMLS concept id:\textit{C0024117} has the unique name Chronic Obstructive Airway Disease. This name is unique in UMLS. If we find a text span with this name we can use the surrounding text of this span for training, because it uniquely links to C0024117. $\sim95\%$ of the concepts in UMLS have at least one unique name.

The context of a concept is represented by vector embeddings. Given a document $d_1$ where $C_x$ is a detected concept candidate (Equation. \ref{eq:doc}) we calculate the context embedding. This is a vector representation of the context for that concept candidate (Equation. \ref{eq:calc_cntx}). That includes a pre-set (s) number of words to the left and right of the concept candidate words. Importantly, the concept candidate words are also included in context embedding calculation as the model is assisted by knowing what words the surrounding context words relate to.

\begin{align}
    d_1 = w_1\ w_2\ \cdots\ \overbrace{w_k\ w_{k+1}}^{C{x}}\ \cdots\ w_n  \label{eq:doc}
\end{align}
Where:\\
\hspace*{2mm} $d_1$ - An example of a document \\
\hspace*{2mm} $w_{1..n}$ - Words in the document, or to be more specific tokens \\
\hspace*{2mm} $C_x$ - The detected concept candidate that matches the words $w_k$ and $w_{k+1}$

\begin{align}
V_{cntx} = \frac{1}{2s} \Big[\sum_{i=1}^s V_{w_{k-i}} + \sum_{i=1}^s V_{w_{k+1+i}}] \label{eq:calc_cntx}
\end{align}
Where:\\
\hspace*{2mm} $V_{cntx}$ - Calculated context embedding \\
\hspace*{2mm} $V_{w_k}$ - Word embedding \\
\hspace*{2mm} $s$ - Words from left and right that are included in the context of a detected concept candidate. Typically, $s$ is set to $9$ for \textit{long} context and $2$ for \textit{short} context.

To calculate context embeddings we use the word embedding method Word2Vec\cite{Mikolov2013-sh}. Contextualised embedding approaches such as BERT\cite{Devlin2018-qg} were also tested alongside fastText\cite{Bojanowski2017-ua} and GloVe\cite{Pennington2014-fn}. Results presented in Section \ref{sec:res_nerl} show the BERT embeddings (the MedCAT U/MI/B configuration) perform worse on average compared to the simpler Word2Vec embeddings. FastText and GloVe perform similarly to Word2Vec, therefore our default implementation uses Word2Vec for ease of implementation. We trained 300 dimensional Word2Vec embeddings using the entire MIMIC-III\cite{Johnson2016-mq} dataset of 53,423 admissions. 

Once a correct annotation is found (a word uniquely links to a CDB name), a context embedding $V_{cntx}$ is calculated, and the corresponding $V_{concept}$ is updated using the following formula: 

\begin{align}
    \label{eq:update_pos}
    sim =&\ max(0, \frac{V_{concept}}{\lVert{}V_{concept}\rVert{}} \cdot \frac{V_{cntx}}{\lVert{}V_{cntx}\rVert{}}) \\
    lr =&\ \frac{1}{C_{concept}} \\
    V_{concept} =& V_{concept} + lr \cdot (1 - sim) \cdot V_{cntx}
\end{align}

Where:\\
\hspace*{2mm} $C_{concept}$ - Number of times this concept appeared during training \\
\hspace*{2mm} $sim$ - Similarity between $V_{concept}$ and $V_{cntx}$ \\
\hspace*{2mm} $lr$ - Learning rate

The update rule is based on the Word2Vec model and aims to make the concept embedding $V_{concept}$ similar to the context in which the concept was presently found $V_{cntx}$. The scaling which is achieved via the cosine similarity is used to favour new contexts in which a concept appears over contexts that frequently appeared in the past.

To prevent the context embedding for each concept being dominated by most frequent words, we used negative sampling as defined in\cite{Mikolov2013-sh}. Whenever we update the $V_{concept}$ with $V_{cntx}$ we also generate a negative context by randomly choosing $K$ words from the vocabulary consisting of all words in our dataset. Here $K$ is equal to $2s$ i.e. twice the window size for the context ($s$ is the context size from one side of the detected concept, meaning in the positive cycle we will have s words from the left and $s$ words from the right). The probability of choosing each word and the update function for vector embeddings is defined as:

\begin{align}
    P(w_{i}) =& \frac{f(w_i)^{3/4}}{\sum_j^n f(w_j)^{3/4}} \\
    f(w_i) =& \frac{C_{w_i}}{\sum_j^n C_{w_j}} \\
    V_{ncntx} =&\ \frac{1}{K} \sum_{i}^K V_{w_i} \\
    sim =&\ max(0, \frac{V_{concept}}{\lVert{}V_{concept}\rVert{}}  \cdot
    \frac{V_{ncntx}}{\lVert{}V_{ncntx}\rVert{}}) \\
    V_{concept} =& V_{concept} - lr \cdot sim \cdot V_{ncntx}
    \label{eq:concept_update}
\end{align}

Where: \\
\hspace*{2mm} $n$ - Size of the vocabulary \\
\hspace*{2mm} $P(w_{i})$ - Probability of choosing the word $w_i$ \\
\hspace*{2mm} $K$ - Number of randomly chosen words for the negative context \\
\hspace*{2mm} $V_{ncntx}$ - Negative context

\subsubsection{Supervised Training Procedure}
The supervised training process is similar to the self-supervised process but given the correct concept for the extracted term we update the $V_{concept}$ using the calculated $V_{ctx}$  as defined in Eq. \ref{eq:update_pos}-\ref{eq:concept_update}. This no longer relies upon the self-supervised constraint that at least one name in the set of possible names for a concept is unique as the correct term is provided by human annotators. 

\subsubsection{Contextualisation of Identified and Linked Concepts: Meta-Annotations}
Once a span of text is recognised and linked to a concept, further contextualisation or meta-annotation is often required. For example, a simple task of identifying all patients with a fever can entail classifying the located fever text spans that are current mentions (e.g. the patient reports a fever vs the patient reported a fever but \ldots{}), are positive mentions (e.g. patient has a high fever vs patient has no sign of fever), are actual mentions (e.g. patient is feverish vs monitoring needed if fever reappears), or are experienced by the patient (e.g. pts family all had high fevers). We treat each of these contextualization tasks as distinct binary or multiclass classification tasks Meta-annotations are equivalent to `attributes' in cTakes parlance.

The MedCAT library provides a ‘MetaCAT’ component that wraps a Bidirectional-Long-Short-Term-Memory (Bi-LSTM) model trainable directly from MedCATtrainer project exports. Bi-LSTM models have consistently demonstrated strong performance in biomedical text classification task\cite{Luo2018-fk,Wang2019-rg,Xu2018-sw} and our own recent work\cite{Mascio2020-bz} demonstrated a Bi-LSTM based model outperforms all other assessed approaches, including Transformer models. MetaCAT models replace the specific concept of interest for example ‘diabetes mellitus’ with a generic parent term of the concept ‘[concept]’. The forward / backward pass of the model then learns a concept agnostic context representation of the concept allowing MetaCAT models to be used across concepts as observed in our results (Section. \ref{sec:res_meta}). The MetaCAT API follows standard neural network training methods but are abstracted away from end users whilst still maintaining enough visibility for users to understand when MetaCAT models have been trained effectively. Each training epoch displays training and test set loss and metrics such as precision, recall and F1. An open-source tutorial showcasing the MetaCAT features are available as part of the series of wider MedCAT tutorials\footnote{\url{https://colab.research.google.com/drive/1zzV3XzFJ9ihhCJ680DaQV2QZ5XnHa06X}}. Once trained, MetaCAT models can be exported and reused for further usage outside of initial classification tasks similarly to the MedCAT NER+L models.

\subsection{MedCATTrainer: Annotation Tool}
MedCATtrainer allows domain experts to inspect, modify and improve a configured MedCAT NER+L model. The tool either actively trains the underlying model after each reviewed document (facilitating live model improvements as feedback is provided by human users) or simply collects and validates concepts extracted by a static MedCAT model. The active learning is done on a concept level and MedCATtrainer will automatically mark some concepts as correct/incorrect and ask for user input for others where it is not confident enough. Version 0.1\cite{Searle2019-rq} presented a proof-of-concept annotation tool that has been rewritten and tightly integrated with the MedCAT library, whilst providing a wealth of new features supporting clinical informatics workflows. We also provide extensive documentation \footnote{\url{https://github.com/CogStack/MedCATtrainer/blob/master/README.md}} and pre-built containers \footnote{\url{https://hub.docker.com/r/cogstacksystems/medcat-trainer}} updated with each new release facilitating easy setup by informatics teams.

\subsection{Datasets and Experimental Setup}\label{sec:ds_and_exp}

\subsubsection{Named Entity Recognition and Linking Open Datasets}
MedCAT concept recognition and linking was validated on the following publicly datasets:
\begin{enumerate}[leftmargin=*]
    \item MedMentions\cite{Mohan2019-wf} - consists of 4,392 titles and abstracts randomly selected from papers released on PubMed in 2016 in the biomedical field, published in the English language, and with both a Title and Abstract. The text was manually annotated for UMLS concepts resulting in 352,496 mentions. We calculate that $\sim40\%$ of concepts in MedMentions require disambiguation, suggesting a detected span of text can be linked to multiple UMLS concepts if only the span of text is considered.
    \item ShARe/CLEF 2014 Task 2\cite{Mowery2014-ht} - we used the development set containing 300 documents of 4 types - discharge summaries, radiology, electrocardiograms, and echocardiograms. We’ve used the UMLS annotations and ignored the attribute annotations.
    \item MIMIC-III\cite{Johnson2016-mq} - consists of $\sim58,000$ de-identified EHRs from critical care patients collected between 2001-2012. MIMIC-III includes demographic, vital sign, and laboratory test data alongside unstructured free-text notes.
\end{enumerate}

We attempted to use the SemEval 2019 shared task for the evaluation of the NER+L task\footnote{\url{https://competitions.codalab.org/competitions/19350}}, but dataset access is currently under review for all requests to i2b2. 

\subsubsection{Clinical Use Case Datasets}\label{sec:clnc_datasets}

Our further experiments used real world EHR data from the following UK NHS hospital Trusts:

\begin{itemize}[leftmargin=*]
    \item King’s College Hospital Foundation Trust (KCH) Dataset: 
    \begin{itemize}
        \item 300 free text inpatient notes for Covid-19 positive patients, 121 Epilepsy clinic letters 2018-2019, 100 Cardiac Clinic letters, 200 echocardiographic reports, 100 CT pulmonary angiograms, 700 10k character chunks of clinical notes of patients with Diabetes Mellitus/ Gastroenteritis/ Inflammatory bowel disease/ Crohn's disease/ Ulcerative colitis for supervised training.
        \item $\sim17M$ documents with $\sim8.8B$ tokens (entire KCH electronic health record from 1999 to 2020 consisting documents from ‘multi-era’, multi-vendor electronic health records (including iSoft iCM, EMIS Symphony and AllScripts) and multiple geographically-distributed hospital sites (Kings College Hospital, Princess Royal University Hospital and Orpington Hospital) were processed for self-supervised training.
    \end{itemize}
    \item South London and Maudsley Foundation Trust (SLaM): 2200 free text notes for patients with a primary or secondary diagnosis of severe mental illness between 2007 and 2018 with each document reviewed for only a specific physical health comorbidity that may or may not appear in the note.
    \item University College London Hospitals Foundation Trust (UCLH) Covid-19 Datasets: 300 Free text clinical notes for Covid-19 positive or suspected patients from Jan - Apr 2020 from single-vendor electronic health record (Epic).
\end{itemize}

We used two large biomedical concept databases and prepared them as described in our source-code repository\footnote{\url{https://github.com/CogStack/MedCAT\#building-concept-databases-from-scratch}}, the databases are:
\begin{itemize}
    \item UMLS 2018AB: 3.82 million concepts and 14 million unique concept names from 207 source vocabularies.
    \item SNOMED CT UK edition: $>$659K concepts. The UK SNOMED CT clinical extension 20200401 and UK Drug Extension 20200325 with ICD-10 and OPCS-4 mappings.
\end{itemize}

\subsubsection{Named Entity Recognition and Linking Experimental Setup}
We use MedMentions\cite{Mohan2019-wf}, ShARe/CLEF\cite{Mowery2014-ht} and MIMIC-III\cite{Johnson2016-mq} datasets in our experiments. We denote the ‘MedMentions’ dataset (i.e. all concepts) and ‘MedMentions Disorders Only’ (i.e. only concepts grouped under the Disorder group as shown in\cite{pmid14759816}). We train MedCAT self-supervised on MIMIC-III configured with the UMLS database. We denote the version using Word2Vec embeddings as ‘MedCAT’ and the one using Bio\_ClinicalBERT\cite{Alsentzer2019-ji} embeddings as ‘MedCAT BERT’.

An annotation by MedCAT is considered correct only if the exact text value was found and the annotation was linked to the correct concept in the CDB. We contrast our performance with the performance of tools presented in Section. \ref{sec:existing_tools}. \ref{appdx:training_config} provides self-supervised training configuration details.

\subsubsection{Clinical Use Case NER+L Experimental Setup}\label{sec:clnc_exp_setup}
For our clinical use cases we extracted SNOMED-CT terms, the official terminology across primary and secondary care for the UK National Health Health service, as this was preferred by our clinical teams over UMLS. 

\begin{figure*}
    \centering
    \includegraphics[scale=0.35]{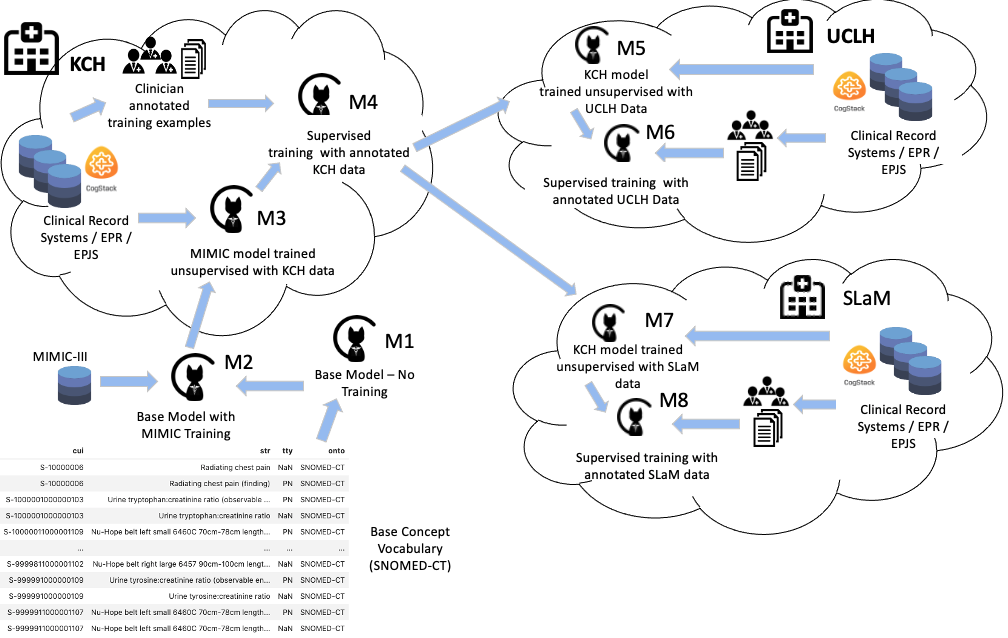}
    \caption{Model provenance for NER+L clinical use case results between datasets and sites. M1-8, showing the MedCAT model instances, the data and method of training and base model used across all sites.}
    \label{fig:medcat_model_provenance}
\end{figure*}

Fig. \ref{fig:medcat_model_provenance} shows our process of model training and distribution to partner hospital Trusts. Initially, we built our untrained MedCAT model using the SNOMED-CT concept vocabulary (M1), we then trained it self-supervised on the MIMIC-III dataset (M2). Next, the entire KCH EPR (17M documents with 8.8B tokens) is used for self-supervised training (M3). We collect annotations with clinician experts at KCH and train supervised (M4). We share this model with each partner hospital site where further self-supervised training (M5, M7) and specific supervised training with their respective annotation datasets (M6, M8). 

Site-specific models (M3, M5, M7) are loaded into deployed instances of MedCATtrainer and configured with annotation projects to collect  SNOMED-CT annotations for a range of site specific disorders, findings, symptoms, procedures and medications that our clinical teams are interested in for further research (i.e. already published work on Covid-19[5,6]). These included chronic (i.e. diabetes mellitus, ischaemic heart disease, heart failure) and acute (cerebrovascular accident, transient ischemic attack) disorders. For comparison between sites we find 14 common extracted concept groups (\ref{tab:appdx_concept_table}) and calculate F1 scores for each concept group and reporting average, standard deviation (SD), and interquartile-range (IQR).

We shared fine-tuned MedCAT models between KCH and 2 NHS partner Trusts UCLH and SLaM. This was a collaborative effort with each hospital team only having access to their respective hospital EHR / CogStack instance. Each site collected annotated data using MedCATtrainer, tested the original base model, a self-supervised only trained model and a final supervised trained model with the MedCATtrainer collected annotations. 

\subsubsection{Clinical Use Case Contextualisation Model Experimental Setup}
From ongoing and published work\cite{Bean2020-vq,Carr2020-tj} we configured and collected meta-annotation training examples and trained a variety of contextualisation models per site as defined in Table. \ref{tab:meta_anno_defs}.

\begin{table}[]
    \centering
    \begin{tabular}{p{1cm} p{2.5cm} p{3cm}}
        \toprule
        \textbf{Site} & \textbf{Task} & \textbf{Values}\\
        \hline
        KCH & Presence & Affirmed / Negated / Hypothetical \\
            & Experiencer & Patient / Family / Other \\
            & Temporality & Past / Present / Future \\
        \hline
        UCLH & Negation & Yes / No \\
             & Experiencer & Yes / No \\ 
             & Problem Temporality & Past Medical Issue / Current Problem \\ 
             & Certainty & Confirmed / Suspected  \\
             & Irrelevant & Yes / No \\
         \hline
        SLaM & Status & Patient / Other / NA \\
             & Diagnosis & Yes / No \\
         \bottomrule
    \end{tabular}
    \caption{Meta Annotation Tasks Defined Per Site, KCH = King's College Hospital NHS Foundation Trust, UCLH = University College London Hospitals NHS Foundation Trust, SLaM = South London and Maudsley NHS Foundation Trust}
    \label{tab:meta_anno_defs}
\end{table}

Our experiments test the effectiveness of our meta annotation modelling approach to flexibly learn contextual cues by assessing cross-disorder and cross-site transferability (Section. \ref{sec:res_meta}. To assess cross-disorder transferability of each of the 11 disorder groups (as specified in \ref{tab:appdx_concept_table}) we use the SLaM collected ‘Diagnosis’ dataset that consists of ~100 annotations for each disorder group. We stratify our train/test sets by disorder, placing all examples for one disorder group in the test set and use the remaining disorder examples as a train set. We run this procedure 11 times so that each disorder group is tested once. We average all scores of each fold and report results. 

To demonstrate cross-site transferability we derive an equivalent meta-annotation dataset from the ‘Presence’ (KCH) and ‘Status’ (SLaM) datasets as they are semantically equivalent despite having different possible annotation values. We merge ‘Presence’ annotations from Affirmed/Hypothetical/False to Affirmed/Other to match classes available in SLaM. We then train and test new meta annotation models between sites and datasets report average results. 

\section{Results}\label{sec:results}
We firstly present our concept recognition and linking results, comparing performance across previously described tools in Section. \ref{sec:existing_tools} using the UMLS concept database and openly available datasets presented in Section. \ref{sec:ds_and_exp} We then present a qualitative analysis of learnt concept embeddings demonstrating the captured semantics of MedCAT concepts. Finally, we show real world clinical usage of the deployed platform to extract, link and contextualise SNOMED-CT concepts across multiple NHS hospital trusts in the UK.

\subsection{Entity Extraction and Linking}\label{sec:res_nerl}
Table \ref{tab:nerl_results} presents our results for self-supervised training of MedCAT and NER+L performance compared with prior tools using openly available datasets. Metrics for all the tools were calculated consistently. Bold indicates best performance. For each manual annotation we check whether it was detected and linked to the correct Unified Medical Language System (UMLS) concept. The metrics are precision (P), recall (R) and the harmonic mean of precision and recall (F1). MedCAT models were configured with UMLS concepts and trained (self-supervised) on MIMIC-III: the base version (MedCAT) uses Word2Vec embeddings (trained on MIMIC-III), while (MedCAT BERT) uses static word embeddings from Bio\_ClinicalBERT\cite{Alsentzer2019-ji}. For the BERT version of MedCAT we do not use the full BERT model to calculate context representations, but only the pre-trained static word embeddings.

\begin{table*}[]
\centering
\begin{tabular}{llllllllll} \toprule
\textbf{Model} \textbackslash \textbf{Dataset} &
  \multicolumn{3}{c}{\textbf{MedMentions}} &
  \multicolumn{3}{c}{\textbf{MedMentions (Disorders Only)}} &
  \multicolumn{3}{c}{\textbf{ShARe/CLEF}} \\
  \hline
    & P       & R       & F1      & P       & R       & F1     & P       & R      & F1     \\
SemEHR            & 0.252   & 0.165   & 0.200   & 0.295   & 0.499   & 0.371  & 0.680   & 0.623  & 0.650  \\
Bio-YODIE         & 0.316   & 0.143   & 0.197   & 0.445   & 0.366   & 0.402  & 0.700   & 0.607  & 0.650  \\
cTAKES            & 0.284   & 0.129   & 0.178   & 0.313   & 0.375   & 0.342  & 0.567   & 0.640  & 0.601  \\
MetaMap           & 0.305   & 0.465   & 0.368   & 0.358   & 0.460   & 0.403  & 0.755   & 0.540  & 0.630  \\
ScispaCy*         & \textbf{0.451}   & 0.408   & 0.429   & 0.487   & 0.443   & 0.464  & 0.711   & 0.463  & 0.561  \\
CLAMP*            & 0.324   & 0.067   & 0.110   & \textbf{0.533}   & 0.236   & 0.327  & 0.772   & 0.447  & 0.566  \\
MedCAT BERT     & 0.386   & 0.475   & 0.426   & 0.459   & 0.513   & 0.485  & 0.788   & 0.678  & 0.729  \\
MedCAT            & 0.406   & \textbf{0.500}   & \textbf{0.448}   & 0.470   & \textbf{0.523}   & \textbf{0.495}  & \textbf{0.796}   & \textbf{0.688}  & \textbf{0.738}  \\
\hline
+ $\delta$ (MedCAT-Best) & -0.045   & 0.035   & 0.019   & -0.063   & 0.024   & 0.031  & 0.041   & 0.048  & 0.088  \\
\bottomrule 
\end{tabular}
\caption{Comparison of NER+L tools for the extraction of UMLS concepts. *The results for ScispaCy/CLAMP are not directly comparable to other tools as they are supervised models.}
\label{tab:nerl_results}
\end{table*}

Our results show MedCAT improves performance compared to all prior tools across all tested metrics (excluding precision when compared to ScispaCy/CLAMP - which are supervised models). We observe that the best performance across all tools is achieved on the ShARe/CLEF dataset. However, MedCAT still improves F1 performance by ~9 percentage points over the next best system. We note the simpler Word2Vec embedding (base MedCAT) on average performs better than the more expressive Bio\_ClinicalBERT (BERT) embeddings. We provide a further breakdown of the range of performances by MedCAT across MedMentions and ShARe/CLEF split by UMLS semantic type in Table. \ref{tab:medcat_sem_types}.

\begin{table*}[]
    \centering
    \begin{tabular}{c c c c c c c c}
        \toprule
        \multirow{2}{*}{\textbf{Semantic Type}} & \multirow{2}{*}{\textbf{Dataset}} & \multicolumn{3}{c}{\textbf{MedMentions}} & \multicolumn{3}{c}{\textbf{ShARe/CLEF}} \\
        & & \textbf{P} & \textbf{R} & \textbf{F1} &  \textbf{P} & \textbf{R} & \textbf{F1} \\\midrule
        T047 & Disease or Syndrome & 0.59 & 0.59 & 0.59 & 0.87 & 0.75 & 0.80 \\
        T121 &  Therapeutic or Preventive Procedure & 0.52 & 0.52 & 0.52 & \multicolumn{3}{c}{NO DATA} \\
        T061 &  Pharmacologic Substance & 0.49 & 0.38 & 0.43 & \multicolumn{3}{c}{NO DATA} \\
        T184 &  Sign or Symptom & 0.58& 0.70& 0.64 & 0.86 & 0.75 & 0.80 \\
        T048 & Mental or Behavioral Dysfunction & 0.63 & 0.55 & 0.58 & 0.71 & 0.63 & 0.66 \\
         \bottomrule
    \end{tabular}
    \caption{MedCAT performance for different UMLS semantic types on MedMentions and ShARe/CLEF}
    \label{tab:medcat_sem_types}
\end{table*}

\color{black}

\subsection{Qualitative Analysis}\label{sec:qual_analysis}
For concept disambiguation the MedCAT core library learns vector embeddings from the contexts in which a concept appears. This is similar to prior work\cite{Beam2018-vr}, although we also present a novel self-supervised training algorithm, annotation system and wider workflow. Using our learnt concept embeddings we perform a qualitative analysis by inspecting concept similarities, with the expectation that similar concepts have similar embeddings. Table. \ref{tab:qual_analysis} shows the learnt context embeddings capture medical knowledge including relations between diseases, medications and symptoms. We train MedCAT self-supervised over MIMIC-III\cite{Johnson2016-mq} using the entirety of UMLS, 3.82 Million concepts from 207 separate vocabularies. Training configuration details are provided in \ref{appdx:training_config}.

\setlength\tabcolsep{2pt}
\begin{table*}[]
\centering
    \begin{tabular}{lll}
    \toprule
    Disease $\rightarrow{}$ Medication & Disease $\rightarrow{}$ Procedure & Symptom  $\rightarrow{}$ Medication \\
    \hline
    \textbf{Hypertensive disease}             & \textbf{Neoplastic Process}     & \textbf{Fever} \\
    \hline
    Metoprolol 50 MG                 & Chemotherapy                    & Levofloxacin\\
    Metoprolol 25 MG                 & Radiosurgery                    & Vancomycin\\
    Valsartan 320 MG                 & FOLFOX Regimen                  & Vancomycin 750 MG\\
    Nadolol 20 MG                    & Chemotherapy Regimen            & Azithromycin\\
    Atenolol 100 MG                  & Preoperative Therapy            & Levofloxacin 750 MG\\
    Enalapril 10 MG                  & Anticancer therapy              & Dexamethasone\\
    Oral form diltiazem              & Parotidectomy                   & Lorazepam \\
    nimodipine 30 MG                 & Resection of ileum              & Acetaminophen\\
    \bottomrule
    \end{tabular}
\caption{Qualitative Analysis of Learnt Concept Embeddings. UMLS concepts that have highest cosine similarity between learnt vector embeddings of concepts in \textbf{bold}. The first row defines the chosen concept and the target concept type. We have randomly chosen the most frequent concepts and presented the 8 most similar concepts for each target concept type. For example, Neoplastic Process (C0006826) and the following rows show the top 8 most similar Procedure concepts.}
\label{tab:qual_analysis}
\end{table*}

\subsection{Clinical Use Cases across Multiple Hospitals}\label{sec:clnc_cases}
The MedCAT platform was used in a number of clinical use cases providing evidence for its applicability to answer relevant, data intensive research questions. For example, we extracted relevant comorbid health conditions in individuals with severe mental illness and patients hospitalized after Covid-19 infection\cite{Bean2020-vq,Carr2020-tj,Zakeri2020-zg}. These use cases analysed data sources from 2 acute secondary/tertiary care services at King's College Hospital (KCH), University College London Hospitals (UCLH) and mental health care services South London and Maudsley (SLaM) NHS Foundation Trusts in London, UK.

The following results focus on providing an aggregate view of MedCAT performance over real NER+L clinical use-cases, meta-annotation or context classification tasks and model transferability across clinical domains (physical health vs mental health), EHR systems and concepts.

\subsubsection{Entity Extraction and Linking}
Table. \ref{tab:res_clini_nerl} shows our results for NER+L across hospital sites, model and training configurations as described in Section \ref{sec:clnc_datasets} Our KCH annotations were collected across a range of clinicians, clinical research questions and therefore MedCATtrainer projects. This unfortunately led to a lack of resourcing to enable double annotations and calculation of inter-annotator-agreement (IIA) scores. SLaM annotations were collected by clinician / non-clinician pairs with average inter-annotator agreement (IIA) at 0.88, disagreements were discarded before results were calculated to ensure a gold-standard. UCLH IIA was at 0.85 between two medical students with annotation disagreements arbitrated by an experienced clinician providing the final gold-standard dataset. For our KCH results we use all annotations collected across various MedCATtrainer projects within our 14 concept groups as described in Section. \ref{sec:clnc_exp_setup} Both KCH and UCLH annotations contained occurrences of all 14 concept groups, SLaM annotated notes did not contain any occurrences of Dyspnea (SCTID:267036007), Pulmonary embolism (SCTID:59282003) and Chest pain (SCTID:29857009).

\setlength\tabcolsep{5pt}
\begin{table*}
\centering
\begin{tabular}{p{1cm} p{6cm} p{2cm} p{2cm} p{1cm} p{1cm} p{1cm}}
    \toprule
    \textbf{Model} & \textbf{Training Configuration} & \textbf{Hospital Test Site} & \textbf{\# Annotated Examples} & \textbf{F1 $\mu$}  & \textbf{F1 SD$\pm$} & \textbf{F1 IQR} \\
    \midrule
     M1 & Base - No Training                                                          & KCH   & 3,358 & 0.638 & 0.297  & 0.333 \\
    \hline
    M2 & Base + Self-Supervised MIMIC-III                            & KCH & 3,358 & 0.840 & 0.109  & 0.150 \\
    \hline
    M3 & Base + Self-Supervised KCH                                   & KCH & 3,358 & 0.889 & 0.078  & 0.103 \\
    \hline
    M4 & KCH Self-Supervised + KCH Supervised                         & KCH & 3,358 & 0.947 & 0.044  & 0.051 \\
    \hline
    M4 & KCH Self-Supervised + KCH Supervised      & UCLH & 499  & 0.903 & 0.103  & 0.112 \\
    \hline
    M5 & KCH Self-Supervised + KCH Supervised + UCLH Self-Supervised  & UCLH & 499  & 0.905 & 0.079  & 0.034  \\
    \hline
    M6 & KCH Self-Supervised + KCH Supervised + UCLH Self-Supervised + UCLH Supervised & UCLH & 499                   & 0.926 & 0.060  & 0.086  \\
    \hline
    M4 & KCH Self-Supervised + KCH Supervised & SLaM & 1,425 & 0.885 & 0.095  & 0.088  \\
    \hline
    M7 & KCH Self-Supervised + KCH Supervised + SLaM Self-Supervised  & SLaM & 1,425 & 0.907 & 0.047  & 0.082  \\
    \hline
    M8 & KCH Self-Supervised + KCH Supervised + SLaM Self-Supervised + SLaM Supervised & SLaM               & 1,425                 & 0.945 & 0.029  & 0.025 \\
    \bottomrule
\end{tabular}
\caption{NER+L Results Across Hospitals. MedCAT NER+L performance for common disorder concepts defined in \ref{tab:appdx_concept_table} by clinical teams. Annotations for supervised learning are used as test sets for models M1, M2, M3, M5, M7. Average performance on a 10 fold cross-validation with a held out test set is reported for models M4, M6, M8. KCH: Kings College Hospital; UCLH: University College Hospital; SLaM: South London and The Maudsley NHS Foundation Trusts.}
\label{tab:res_clini_nerl}
\end{table*}

\subsubsection{Entity Extraction and Linking Model Transferability}
Table. \ref{tab:res_clini_nerl} demonstrates the improved NER+L performance that arises from using domain specific data first self-supervised in MIMIC-III, then KCH. We observe further improvements with clinician expertise with supervised training using the KCH data. With model sharing to UCLH we observe a 0.044 average drop in F1 performance compared to KCH. Further self-supervised training directly on UCLH data offers minimal average performance gains but does reduce the F1 SD and IQR suggesting there is less variability in performance across concepts. Supervised training on a small (499) annotations from UCLH delivers comparable performance to our KCH trained model. For our experiments at SLaM we see average F1 performance drop initially by 0.062 using the KCH model directly on SLaM data. SLaM is a large mental health service provider where EHRs are markedly different to acute care hospitals KCH and UCLH. Interestingly, successive self-supervised (M7) and supervised training (M8) show benefits across all measures with final performance largely similar to final KCH performance. 

Importantly, this suggests performance is transferred to the different hospital sites and initially only drops by $\sim$0.04. With self-supervised training and further supervised training we are able to reach KCH performance with $\sim7\times$ fewer manually collected examples at UCLH or $\sim2\times$ fewer examples at SLaM.

\subsubsection{Contextualisation Model Performance}\label{sec:res_meta}
Contextualisation of extracted and linked concepts is, by design, bespoke per project. Due to this, reporting and comparing results across studies / sites is difficult as the definitions of tasks and concepts collected are different and therefore output trained models are bespoke. Table. \ref{tab:meta_annos_result_a} shows aggregate performance at each site, and Table. \ref{tab:meta_annos_result_b}-\ref{tab:meta_annos_result_c} show further experiments for cross-site and cross-concept model transferability.

We achieve strong weighted (0.892-0.977) / macro (0.841-0.860) F1 performance across all tasks and sites, with breakdown of each metric per site/task available in \ref{appdx:meta_annotation_res}. We report average macro and weighted F1 score demonstrating the variation in performance due to unbalanced datasets across most tasks. 

\setlength\tabcolsep{6pt}
\begin{table*}[]
    \caption{Contextualisation Model Results}
    \begin{subtable}[h]{\textwidth}
        \centering
        \begin{tabular}{lllll}
        \toprule
        \textbf{Site} & \textbf{Task} & \textbf{\# Annotated examples} & \textbf{Macro F1} & \textbf{Weighted F1} \\
        \midrule
        KCH  & Presence & 37,310 & 0.846 & 0.929 \\
         & Temporality & 18,670 & 0.803 & 0.943 \\
         & Experiencer & 18,670 & 0.867 & 0.959\\
        \hline
        SLaM & Patient Diagnosis & 1,152 & 0.904 & 0.913 \\
             & Status  & 1,152 & 0.775 & 0.812 \\
        \hline
        UCLH & Negation & 4,400 & 0.836 & 0.970 \\
            & Experiencer & 4,400 & 0.940 & 0.996 \\
            & Problem Temporality & 4,350 & 0.848 & 0.970 \\  
            & Certainty & 4,160 & 0.836 & 0.970 \\
            & Irrelevant & 4,390 & 0.835 & 0.969 \\ 
        \bottomrule
        \end{tabular}
        \caption{Site Specific Contextualisation Model Performance. Weighted / Macro average F1 Meta annotation model performance custom defined and trained per site - detailed definitions are provided in \ref{appdx:meta_annotation_res}. Task definitions are uniquely defined at each site, e.g. Experiencer at KCH considers the values patient / family / other whereas Experiencer at UCLH only considers the value patient / other. Status at SLaM considers the values affirmed / other and Certainty at UCLH considers the values confirmed / suspected. We include all concepts of interest as defined under clinician guidance at each site, therefore site-to-site comparison in performance cannot be made.}
        \label{tab:meta_annos_result_a}
    \end{subtable}
    \newline
    \vspace*{0.25cm}
    \newline
    \begin{subtable}[h]{\textwidth}
        \centering
        \begin{tabular}{lllll}
        \toprule
        \textbf{Site} & \textbf{Task}      & \textbf{Train / Test Split} & \textbf{Macro F1} &
        \textbf{Weighted F1} \\
        \midrule
        SLaM & Diagnosis & Concept Stratified & 0.82     & 0.85        \\
        SLaM & Diagnosis & Random             & 0.90     & 0.91       \\
        \bottomrule
        \end{tabular}
        \caption{Cross Site Transferability Performance. 11 fold concept stratified CV vs randomized CV for SLaM ‘Diagnosis’ contextualisation task performance. The 11 concepts were selected from NER+L experiment concepts available at SLaM (Supplementary Table 1). The ‘Diagnosis’ task at SLaM was used as this was our most balanced dataset between all tasks and concepts collected.}
        \label{tab:meta_annos_result_b}
    \end{subtable}
    \newline
    \vspace*{0.25cm}
    \newline
    
    \begin{subtable}[h]{\textwidth}
        \centering
        \begin{tabular}{lllll}
        \toprule
        \textbf{Site} & \textbf{Trained on} &
        \textbf{\# Annotated Examples} & \textbf{Macro F1} & \textbf{Weighted F1}\\
        \midrule
        KCH                 & KCH                       & 37,310                & 0.89     & 0.93        \\
        SLaM                & KCH                       & 37,310                 & 0.71     & 0.91        \\
        SLaM                & SLaM                      & 1,152                 & 0.77     & 0.87        \\
        SLaM                & KCH + SLaM & 38,462 & 0.85     & 0.96 \\
        \bottomrule
        \end{tabular}
        \caption{Cross-site transferability of the MetaCAT model for Presence (at KCH) / Status (at SLaM converted to values of Affirmed/Other) - as that was the only task that existed across sites. Results show 10 fold CV where applicable - e.g. row 2 is direct testing of the KCH model on SLaM data, so no training is performed on the SLaM side.}
        \label{tab:meta_annos_result_c}
    \end{subtable}
\end{table*}

For cross-concept transferability, Table. \ref{tab:meta_annos_result_b} shows a decrease in performance when stratifying by concept. However, we still observe a relatively high 0.82-0.85 score suggesting the model is capable of learning disorder independent representations that distinguish the classification boundary for the ‘Diagnosis’ task, not just the disorder specific contexts.

Our cross-site transferability results, Table \ref{tab:meta_annos_result_c}, suggest the ‘Status’ context model that is trained on cross site (i.e. KCH) data then fine-tuned on site specific data (i.e. SLaM) performs better (+ 0. 0.08 Macro / + 0.09 Weighted F1) compared with training on only the SLaM site specific training only (i.e. comparing row 3 and 4). 

\section{Discussion}

\subsection{Named Entity Recognition and Linking}
Our evaluation of MedCAT’s NER+L method using self-supervised training was bench-marked against existing tools that are able to work with large biomedical databases and are not use-case specific. Our datasets and methods are publicly available making the experiments transparent, replicable, and extendable. With the MedMentions dataset, using only self-supervised learning, our results in \ref{sec:res_nerl}, demonstrate an improvement on the prior tools for both disorder detection (F1=0.495 vs 0.464) and general concept detection (F1=0.448 vs. 0.429). We observe all tools perform best with the ShARe/CLEF dataset. We suggest this broadly due to the lack of ambiguity and the more clinical setting allowing alternative systems to also perform reasonably well. 

We now discuss the result between our BERT and regular (Word2Vec) configured MedCAT models. Generally BERT, a deep neural embedding model, performs well for a range of downstream tasks\cite{Devlin2018-qg} better than older approaches such as Word2Vec, i.e. a shallow neural embedding. We believe this due to our use of pre-trained static BERT embeddings that: 1) are not specifically trained to produce similar values for words appearing in a similar context, 2) Sub-word tokenization might be problematic if the tokenizer was trained on a non-medical dataset (no matter whether it was fine-tuned later on MIMIC-III, pubmed or similar).

The general concept detection task with MedMentions is difficult due to: the larger number of entities to be extracted, the rarity of certain concepts and the often highly context dependent nature of some occurrences. Recent work\cite{Fraser2019-mu} highlights examples of ambiguous texts within the MedMentions dataset such as ‘probe’ with 7 possible labels (‘medical device’, ‘indicator reagent or diagnostic aid’ etc.) Further work[40] also showed a deep learning approach (BioBERT+) that achieved F1=0.56. When MedCAT is provided with the same supervised training data we achieve F1=0.71. We find our improved performance is due to the long tail of entities in MedMentions that lack sufficient training data for methods such BioBERT to perform well. 

Our qualitative inspection of the learnt concept embeddings, \ref{sec:qual_analysis} indicate learnt semantics of the target medical domain. This result mirrors similar findings reported in fields such as materials science\cite{Tshitoyan2019-nw}. Recent work has suggested an approach to quantity the effectiveness of learnt embeddings[38] in representing the source ontology. However, this relies on concept relationships to be curated before assessment requiring clinical guidance that may be subjective in the clinical domain. We leave a full quantitative assessment of the learnt embeddings to future work for this reason.

As more concepts are extracted the likelihood of concepts requiring disambiguation increases, particularly in biomedical text\cite{Krauthammer2004-vh}. Estimating the number of training samples for successful disambiguation is difficult but based on our experiments we need at least 30 occurrences of a concept in the free text to perform disambiguation. We provide more details in \ref{appdx:estimating_counts}.

Finally, we note that there are no limitations algorithmically for MedCAT to support languages other than our tested language, English. As MedCAT uses a concept dictionary/vocab for NER+L, if there are existing resources (e.g. SNOMED-CT has already been translated into Spanish, Dutch, Swedish and Danish) they can be used directly for these languages with likely similar results. Alternatively, users could build their own custom concept dictionary (CDB) for their language of choice. Meta-annotation or contextualisation models also do not have language specific features, i.e. English, and would also likely perform well as they only rely on bi-directional context from supervised examples to make predictions.

\subsection{Clinical Use Cases}
MedCAT models and annotated training data have been implemented to be easily shared and reused, facilitating a federated learning approach to model improvement and specialisation with models brought to sensitive data silos. Our results in Section. \ref{sec:clnc_cases} demonstrate that we are able directly apply models trained at one hospital site (KCH) to multiple other sites, and clinical domains (physical vs mental health datasets) with only a small drop in average F1 (0.044 at UCLH, 0.062 at SLaM), and after small amount of additional site specific training, we observe comparable performance (-0.021 at UCLH, -0.002 at SLaM). 

We also highlight that separate teams were able to deploy, extract and analyse real clinical data using the tools as is by following provided examples, documentation and integrations with the wider CogStack ecosystem. Academic engineering projects are often built to support a single research project, however MedCAT and the CogStack ecosystem are scalable fit-for-purpose locally-tunable solutions for teams to derive value from their data instead of being stalled by poor quality code or lack of documentation. This means the model is broadly useful with top-up tuning also available for specific scenarios, domains and hospitals.

Each hospital site and clinical team freely defined the set of contextualisation tasks and associated values for each task. On aggregate our results show performance is consistently strong across all sites and tasks (Macro F1: 0.841-0.860, Weighted F1: 0.892-0.977). With many of the tasks the annotated datasets are highly unbalanced. For example, the ‘Presence’ task at KCH, disorders are often only mentioned in the EHR if they are affirmed (e.g. “...pmhx: TIA...”), and only rarely are hypothetical (e.g. “...patient had possible TIA...”) or negated terms (e.g. “...no sign of TIA…”) encountered. This explains the differences in performance when reporting macro vs weighted average F1 score. We would expect generalization performance to lie between these reported metrics.

\subsection{Limitations}
MedCAT is able to employ a self-supervised training method as the initial pass of the algorithm uses a given unique name to learn and improve an initial concept embedding. However, if the input vocabulary linked to the concepts inadequately specifies possible names or the given names of a concept rarely appear in the text then improvements can only occur during standard supervised learning. The main limitation of our approach is that it greatly depends on the quality of the concept database. Large biomedical concept databases (e.g. UMLS) however have a well specified vocabulary offering many synonyms, acronyms and differing forms of a given concept.

A limitation of our concept embedding approach is if different concepts appear in similar contexts disambiguation and linking to the correct concept can be difficult. For example, ‘OD’ can link to ‘overdose’ or ‘once daily’, both referring to medications with very different implications. We have rarely seen this problem during real-world corpus.
Our approach can also struggle if concepts appear in many varying contexts that are rarely seen or annotated for. With each new context updating the underlying concept embedding this may decrease performance of the embedding. 

Supervised learning requires training data to be consistently labelled. This is a problem in the clinical domain that consists of specialised language that can be open to interpretation. We recommend using detailed annotation guidelines that enumerate ambiguous scenarios for annotators.

\subsection{Future Work}
MedCAT uses a vocabulary based approach to detect entity candidates. Future work could investigate the expansion of such an approach with a supervised learning model like BERT\cite{Devlin2018-qg}. The supervised learning model would then be used for detection of entity candidates that have enough training data and to overcome the challenge of detecting new unseen forms of concept names. The vocabulary based approach would cover cases with insufficient annotated training data or concepts that have few different names (forms). The linking process for both approaches would remain the same self-supervised. 

Our self-supervised training over the $\sim$20 year KCH EHR, as described in Section. \ref{sec:ds_and_exp}, took over two weeks to complete. Future work could improve the training speed by parallelizing this process since concepts in a CDB are mostly independent of one another. Further work could address effective model sharing, allowing subsequent users/sites to benefit from prior work, where only model validation and fine-tuning is required instead of training from scratch. 

Finally, ongoing work aims to extend the MedCAT library to address relation identification and extraction. For example, linking the extracted drug dosage /  frequency with the associated drug concept, or identifying relations between administered procedures and following clinical events. 

\section{Conclusions}
This paper presents MedCAT a multi-domain clinical natural language processing toolkit within a wider ecosystem of open-source technologies namely CogStack. 

The biomedical community is unique in that considerable efforts have produced  comprehensive concept databases such as UMLS and SNOMED-CT amongst many others. MedCAT flexibly leverages these efforts in the extraction of relevant data from a corpus of biomedical documents (e.g. EHRs). Each concept can have one or more equivalent names, such as abbreviations or synonyms. Many of these names are ambiguous between concepts. The MedCAT library is based upon a simple idea: at least one of the names for each concept is unique and given a large enough corpus that name will be used in a number of contexts. As the context is learned from the unique name, when an ambiguous name is later detected, its context is compared to the learnt context, allowing us to find the correct concept to link. By comparing the context similarity we can also calculate confidence scores for a provided linked concept. 

With MedCAT we have built an effective, high performance IE algorithm demonstrating improved performance over prior solutions on open access datasets. We have commoditised the development, deployment and implementation of IE pipelines with supporting technologies MedCATtrainer / MedCATservice supporting the transfer, validation, re-use and fine-tuning of MedCAT models across sites, clinical domains and concept vocabularies. MedCAT deployments are enabled by extensive documentation, examples, APIs and supporting real world clinical use cases outlined in prior published work.

Overall, MedCAT is built to enable clinical research and potential improvements of care delivery by leveraging data in existing clinical text. Currently, MedCAT is deployed in a number of hospitals in the UK in silo or as part of the wider CogStack ecosystem, with wide-ranging use cases to inform clinical decisions with real-time alerting, patient stratification, clinical trial recruitment and clinical coding. The large volume of medical information that is captured solely in free text is now accessible using state-of-the-art healthcare specific NLP. 

\section*{Acknowledgements}
We would like to thank all the clinicians who provided annotation training for MedAT; this includes Rosita Zakeri, Kevin O’Gallagher, Rosemary Barker, David Nicholson Thomas, Rhian Raftopoulos, Pedro Viana, Elisa Bruno, Eugenio Abela, Mark Richardson, Naoko Skiada, Luwaiza Mirza, Natalia Chance, Jaya Chaturvedi, Tao Wang, Matt Solomon, Charlotte Ramsey and James Teo. 

RD's work is supported by 1.National Institute for Health Research (NIHR) Biomedical Research Centre at South London and Maudsley NHS Foundation Trust and King’s College London. 2. Health Data Research UK, which is funded by the UK Medical Research Council, Engineering and Physical Sciences Research Council, Economic and Social Research Council, Department of Health and Social Care (England), Chief Scientist Office of the Scottish Government Health and Social Care Directorates, Health and Social Care Research and Development Division (Welsh Government), Public Health Agency (Northern Ireland), British Heart Foundation and Wellcome Trust. 3. The National Institute for Health Research University College London Hospitals Biomedical Research Centre. 

DMB is funded by a UKRI Innovation Fellowship as part of Health Data Research UK MR/S00310X/1 (\href{https://www.hdruk.ac.uk}{https://www.hdruk.ac.uk}). 

RB is funded in part by grant MR/R016372/1 for the King’s College London MRC Skills Development Fellowship programme funded by the UK Medical Research Council (MRC, https://mrc.ukri.org) and by grant IS-BRC-1215-20018 for the National Institute for Health Research (NIHR, \href{https://www.nihr.ac.uk}{https://www.nihr.ac.uk}) Biomedical Research Centre at South London and Maudsley NHS Foundation Trust and King’s College London. 

ADS is supported by a postdoctoral fellowship from THIS Institute.
AS is supported by a King’s Medical Research Trust studentship.
RS is part-funded by: i) the National Institute for Health Research (NIHR) Biomedical Research Centre at the South London and Maudsley NHS Foundation Trust and King’s College London; ii) a Medical Research Council (MRC) Mental Health Data Pathfinder Award to King’s College London; iii) an NIHR Senior Investigator Award; iv) the National Institute for Health Research (NIHR) Applied Research Collaboration South London (NIHR ARC South London) at King’s College Hospital NHS Foundation Trust.

This paper represents independent research part funded by the National Institute for Health Research (NIHR) Biomedical Research Centre at South London and Maudsley NHS Foundation Trust, The UK Research and Innovation London Medical Imaging \& Artificial Intelligence Centre for Value Based Healthcare (AI4VBH); the National Institute for Health Research (NIHR) Applied Research Collaboration South London (NIHR ARC South London) and King’s College London. The views expressed are those of the author(s) and not necessarily those of the NHS, MRC, NIHR or the Department of Health and Social Care. We thank the patient experts of the KERRI committee, Professor Irene Higginson, Professor Alastair Baker, Professor Jules Wendon, Professor Ajay Shah, Dan Persson and Damian Lewsley for their support.

\section*{Data Availability}
Data for reproduction of experiments for the assessment for the core NER+L in comparison with are available from prior work (MedMentions, ShARe/CLEF 2014 Task 2, MIMIC-III). Due to the confidential nature of free-text data, we are unable to make patient-level data available. Interested readers should contact the authors to discuss feasibility of access of de-identified aggregate data consistent with legal permissions.  

\section*{Code Availability}
All code for running the experiments, the toolkit and integration with wider CogStack deployments are available here:

MedCAT: \url{https://github.com/CogStack/MedCAT}

MedCAT Tutorials/Example Code: \url{https://github.com/CogStack/MedCAT/tree/master/tutorial}

MedCATtrainer: \url{https://github.com/CogStack/MedCATtrainer}

MedCATtrainer Examples: \url{https://github.com/CogStack/MedCATtrainer/tree/master/docs}

MedCATservice: \url{https://github.com/CogStack/MedCATservice}

CogStack: \url{https://github.com/CogStack/CogStack-Pipeline}

\section*{Declaration of Interests}
JTHT received research support and funding from InnovateUK, Bristol-Myers-Squibb, iRhythm Technologies, and holds shares <£5,000 in Glaxo Smithkline and Biogen.

\section*{Data Access Ethics}
NER+L experiments use freely available open-access datasets accessible by data owners.
SNOMED-CT and UMLS licences were obtained by all users at all hospital sites. Site specific ethics is listed below.
KCH: This project operated under London South East Research Ethics Committee approval (reference 18/LO/2048) granted to the King’s Electronic Records Research Interface (KERRI); specific work on research on natural language processing for clinical coding was reviewed with expert patient input on the KERRI committee with Caldicott Guardian oversight. Direct access to patient-level data is not possible due to risk of re-identification, but aggregated de-identified data may be available subject to legal permissions.
UCLH: UCLH is deploying CogStack within its records management infrastructure and is growing its capacity to annotate its clinical records as part of wider work for routine curation. The work at UCLH described here is a service evaluation that represents MedCAT’s annotation of the records. Access to the medical records will not be possible given their confidential nature.
SLaM:  This project was approved by the CRIS Oversight Committee which is responsible for ensuring all research applications comply with ethical and legal guidelines. The 
CRIS system enables access to anonymised electronic patient records for secondary analysis from SLaM and has full ethical approvals. CRIS was developed with extensive involvement from service users and adheres to strict governance frameworks managed by service users. It has passed a robust ethics approval process acutely attentive to the use of patient data. Specifically, this system was approved as a dataset for secondary data analysis on this basis by Oxfordshire Research Ethics Committee C (08/H06060/71). The data is de-identified and used in a data-secure format and all patients have the choice to opt-out of their anonymized data being used. Approval for data access can only be provided from the CRIS Oversight Committee at SLaM. 

\section*{Author Contributions}
ZK, TS, JT, RD, AS, AF conceived the study design

ZK, TS, AS, LR, KN performed data processing and software development

ZK, TS, JT, AS, AM, LZ, ADS performed data validation

RD, JT, RS, ZI, AR, DB, ZI, RB, MPR, ADS, AM  performed critical review

TS, ZK, AS, LR, ZI, RB, DB, AM, RD wrote the manuscript

\appendix
\setcounter{table}{0}
\renewcommand{\thetable}{\Alph{section}.\arabic{table}}

\section*{Appendices}

\section{SNOMED-CT Groupings}\label{tab:appdx_concept_table}
Each group was defined with expert clinical guidance. S-267036007 - Dyspnea (finding), S-59282003 - Pulmonary embolism, (disorder) S-29857009 - Chest pain (finding) do not appear in the SLaM annotations for supervised training.

\onecolumn
\begin{longtable}{p{6cm} p{10cm}}
\caption{SNOMED-CT concept level groupings for clinical use cases}\\

\toprule
\textbf{Container Concept} & \textbf{Concepts}\\
\midrule
\endfirsthead

\multicolumn{2}{c}%
{\bfseries Table continued from previous page} \\
\toprule
\textbf{Container Concept} & \textbf{Concepts} \\
\midrule
\endhead

\multicolumn{2}{r}{{Continued on next page}} \\ 
\endfoot

\bottomrule
\endlastfoot

\multirow{5}{*}{S-73211009 - Diabetes mellitus(disorder)}
 & S-44054006 - Diabetes mellitus type 2 (disorder)\\ 
 & S-46635009 - Diabetes mellitus type 1 (disorder)\\ 
 & S-422088007 - Disorder of nervous system co-occurrent and due to diabetes mellitus (disorder)\\
 &  S-25093002 - Disorder of eye co-occurrent and due to diabetes mellitus (disorder)\\ 
 &  S-73211009 - Diabetes mellitus (disorder) \\
 \hline
\multirow{6}{*}{S-84114007 -Heart failure (disorder)} & S-128404006 - Right heart failure (disorder)\\ 
& S-48447003 - Chronic heart failure (disorder)\\ 
& S-56675007 - Acute heart failure (disorder)\\ 
& S-85232009 - Left heart failure (disorder)\\ 
& S-42343007 - Congestive heart failure (disorder)\\ 
& S-84114007 - Heart failure (disorder) \\
\hline
\multirow{5}{*}{\shortstack[l]{S-414545008 - Ischemic heart \\disease (disorder)}} & S-413439005 - Acute ischemic heart disease (disorder)\\
& S-413838009 - Chronic ischemic heart disease (disorder)\\
& S-194828000 - Angina (disorder)\\ 
& S-22298006 - Myocardial infarction (disorder)\\
& S-414545008 - Ischemic heart disease (disorder)\\
 \hline
\multirow{5}{*}{\shortstack[l]{S-38341003 - Hypertensive disorder,\\ systemic arterial (disorder)}} & S-31992008 - Secondary hypertension (disorder)\\
 & S-48146000 - Diastolic hypertension (disorder)\\ 
 & S-56218007 - Systolic hypertension (disorder)\\ 
 & S-59621000 - Essential hypertension (disorder)\\ 
 & S-38341003 - Hypertensive disorder systemic arterial (disorder)\\
\hline
\multirow{3}{*}{\shortstack[l]{S-13645005 - Chronic obstructive \\lung disease (disorder)}} & S-195951007 - Acute exacerbation of chronic obstructive airways disease (disorder)\\ 
 & S-87433001 - Pulmonary emphysema (disorder)\\ 
 & S-13645005 - Chronic obstructive lung disease (disorder)\\
  \hline
S-195967001 - Asthma (disorder) & S-195967001 - Asthma (disorder) \\
  \hline
\multirow{2}{*}{\shortstack[l]{S-709044004 - Chronic kidney \\disease (disorder)}}
    & S-723190009 - Chronic renal insufficiency (disorder)\\ 
    & S-709044004 - Chronic kidney disease (disorder)\\
  \hline
    \multirow{8}{*}{\shortstack[l]{S-230690007 - Cerebrovascular \\accident (disorder)}} & S-25133001 - Completed stroke (disorder)\\
    & S-371040005 - Thrombotic stroke (disorder)\\ 
    & S-371041009 - Embolic stroke (disorder)\\ 
    & S-413102000 - Infarction of basal ganglia (disorder)\\ 
    & S-422504002 - Ischemic stroke (disorder)\\ 
    & S-723082006 - Silent cerebral infarct (disorder)\\
    & S-1078001000000105 - Haemorrhagic stroke (disorder)\\
    & S-230690007 - Cerebrovascular accident (disorder) \\
  \hline
  S-266257000 - Transient ischemic attack (disorder) &
  S-266257000 - Transient ischemic attack (disorder) \\
  \hline
\multirow{5}{*}{S-84757009 - Epilepsy (disorder)} & S-352818000 - Tonic-clonic  epilepsy (disorder)\\ 
 &   S-19598007 - Generalized epilepsy (disorder)\\ 
 &   S-230456007 - Status epilepticus (disorder)\\ 
 &   S-509341000000107 - Petit-mal epilepsy (disorder)\\ 
 &   S-84757009 - Epilepsy (disorder) \\
  \hline
    S-49436004 - Atrial fibrillation (disorder) & S-49436004 - Atrial fibrillation (disorder) \\
  \hline
S-267036007 - Dyspnea (finding) & S-267036007 - Dyspnea (finding) \\
  \hline
S-59282003 - Pulmonary embolism (disorder) & S-59282003 - Pulmonary embolism (disorder) \\
  \hline
S-29857009 - Chest pain (finding) & S-29857009 - Chest pain (finding) \\

\end{longtable}
\twocolumn

\section{Estimating Example Counts for Sufficient F1 Score}\label{appdx:estimating_counts}

To test the required number of examples to achieve a high enough F1 score, we created a mini-dataset from MedMentions. It contains two concepts: C0018810 (Heart Rate) and C2985465 (Hazard Ratio). Both concepts have a unique name and the ambiguous abbreviation HR that can link to either one. We chose these two concepts, as the abbreviation HR is the most frequent ambiguous concept in MedMentions, given the requirement that it must be ambiguous. Our dataset consists of:

\begin{itemize}
    \item 60 training examples (30 per concept). In each example the full name of the concept was used, see below MedMentions Text Extracts.
    \item 174 test examples, each document contains the ambiguous abbreviation HR, see below MedMentions Text Extracts.
\end{itemize}

\begin{figure*}
    \centering
    \includegraphics[scale=0.5]{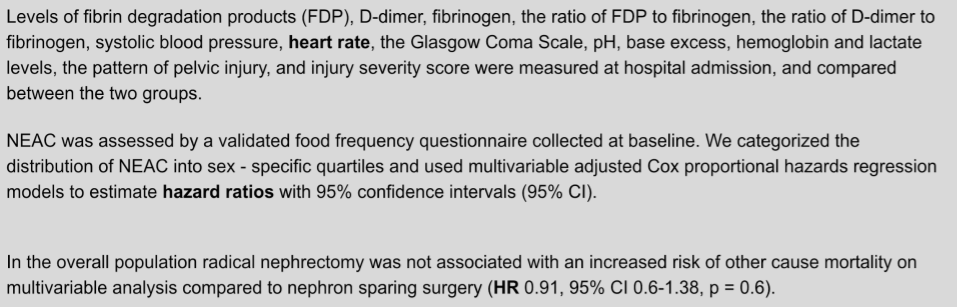}
    \caption{MedMentions Text Extracts: Three samples from the dataset used to test the amount of training samples needed for disambiguation to work. First example is a training case for the concept C0018810, second for C2985465 and third is used to test the disambiguation performance.}
    \label{fig:my_label}
\end{figure*}

We have tested the performance for different sizes of the training set: 1, 5, 10 and 30. If we set the training set size to e.g. 5, we split the full training set into 6 parts (in total the training set has 30 examples per concept), each containing 5 examples per concept. Then we check the performance for each part and report the average over the 6 parts, see Table \ref{tab:examples_for_concept}. 

\begin{table}[]
\begin{tabular}{c c}\toprule
\textbf{Number of examples per concept} & \textbf{F1 on Test} \\\midrule
1                                       & 0.74                \\
5                                       & 0.81                \\
10                                      & 0.82                \\
30                                      & 0.86               \\
\bottomrule
\end{tabular}
\caption{Relation between the number of training examples and performance of MedCAT concept disambiguation.}
\label{tab:examples_for_concept}
\end{table}

\section{Self-Supervised Training Configuration}\label{appdx:training_config}

\subsection{Self-Supervised Training Configuration}
MedCAT was configured for self-supervised training across experiments presented in Section. \ref{medcat_core_lib} as follows:
\begin{itemize}
    \item Misspelled words were fixed only when 1 change away from the correct word for words under 6 characters, and 2 changes away for words above 6 characters.
    \item For each concept we calculate long and short embeddings and take the average of both. The long embedding takes into account s = 9 words from left and right (as shown in Equation 2). The short embedding takes into account s = 2 words from left and right. The exact numbers for s were calculated by testing the performance of all possible combinations for s in the range [0, 10].
    \item The context similarity threshold used for recognition is 0.3 unless otherwise specified. This means for a given concept candidate, or sequence of words, to be recognised and linked to the given concept the concept similarity provided by Equation 2 would be greater than 0.3.
\end{itemize}

\subsection{Qualitative Analysis Training Configuration}
We train MedCAT self-supervised over MIMIC-III using the entirety of UMLS, 3.82 Million concepts from 207 separate vocabularies. We use ~2.4M clinical notes (nursing notes, notes by clinicians, discharge reports etc.) on a small one-core server taking approximately 30 hours to complete.

\section{Contextualisation Task Results Per Site}\label{appdx:meta_annotation_res}

\subsection{Contextualisation Results Breakdown for KCH}
Aggregate results for each defined meta-annotation at KCH. Performance is aggregated over all extracted concepts listed in \ref{tab:appdx_concept_table}. We defined the following meta-annotation tasks: 
\begin{itemize}
    \item Presence: is the concept affirmed, negated or hypothetical, values: [Affirmed, Negated, Hypothetical]
    \item Experiencer: is the concept experienced by the patient or other, values: [Patient / Family / Other]
    \item Temporality: is the concept in the past, present or future, values: [Past, Recent, Future]
\end{itemize}

\setlength\tabcolsep{5pt}
\begin{table}[]
    \caption{Meta Annotation Results at KCH}
    \begin{subtable}[]{0.5\textwidth}
        \begin{tabular}{ccccc}\toprule
        \textbf{CLS} & \textbf{F} & \textbf{P} & \textbf{R} & \textbf{\shortstack[2]{Support Test\\ (10\% of total)}} \\
        \midrule
        Hypothetical & 0.756      & 0.797      & 0.72       & 360                                   \\
        Negated      & 0.865      & 0.878      & 0.852      & 440                                   \\
        Affirmed     & 0.955      & 0.961      & 0.951      & 2930                                  \\
        Macro        & 0.86       & 0.875      & 0.846      & 3731                                  \\
        Weighted     & 0.927      & 0.927      & 0.929      & 3731                                 
        \\\bottomrule
        \end{tabular}
        \caption{Presence average 10 fold CV 90/10 ratio}
    \end{subtable}

    \begin{subtable}[]{0.5\textwidth}
        \begin{tabular}{ccccc}
        \toprule
        \textbf{CLS} & \textbf{F1} & \textbf{P} & \textbf{R} & \textbf{\shortstack[]{Support Test\\(10\% of total)}} \\ \midrule
        Family       & 0.801       & 0.865      & 0.751      & 13                                    \\
        Other        & 0.823       & 0.838      & 0.809      & 205                                   \\
        Patient      & 0.977       & 0.975      & 0.98       & 1649                                  \\
        macro        & 0.867       & 0.893      & 0.847      & 1867                                  \\
        weighted     & 0.959       & 0.959      & 0.959      & 1867                                  \\ \bottomrule
        \end{tabular}
        \caption{Experiencer average 10 fold CV 90/10 ratio}
    \end{subtable}
    \begin{subtable}[]{0.5\textwidth}
        \begin{tabular}{ccccc}\toprule
        \textbf{CLS} & \textbf{F} & \textbf{P} & \textbf{R} & \textbf{\shortstack{Support Test\\(10\% of total)}} \\
        \midrule
        Recent       & 0.969      & 0.964      & 0.94       & 1655                                  \\
        Past         & 0.771      & 0.807      & 0.74       & 162                                   \\
        Future       & 0.667      & 0.706      & 0.74       & 50                                    \\
        macro        & 0.803      & 0.825      & 0.783      & 1867                                  \\
        weighted     & 0.943      & 0.943      & 0.945      & 1867   
                \\\bottomrule
        \end{tabular}
        \caption{Temporality average 10 fold CV 90/10 ratio}
    \end{subtable}
\end{table}

\subsection{Meta Annotation Results Breakdown for SLaM}
Aggregate results for each defined meta-annotation at SLaM. Performance is aggregated over all extracted concepts listed in \ref{tab:appdx_concept_table}. We defined the following meta-annotation tasks: 
\begin{itemize}
    \item Status: is the concept affirmed to be affecting the patient or not, values: [Patient / Other / NA]
    \item Diagnosis: is the concept a diagnosis related to the patient, or not, values: [Yes, No]
\end{itemize}

\begin{table}[]
    \caption{Meta Annotation Results at SLaM}
    \begin{subtable}[]{0.5\textwidth}
        \begin{tabular}{ccccc}
        \toprule
        \textbf{CLS} & \textbf{F} & \textbf{P} & \textbf{R} & \textbf{\shortstack{Support Test\\ (10\% of total)}} \\\midrule
        NA           & 0.873      & 0.869      & 0.878      & 43                                    \\
        Other        & 0.544      & 0.663      & 0.475      & 7                                     \\
        Affirmed     & 0.908      & 0.893      & 0.924      & 60                                    \\
        Macro        & 0.775      & 0.812      & 0.757      & 109                                   \\
        Weighted     & 0.873      & 0.874      & 0.873      & 109                                  \\\bottomrule
        \end{tabular}
        \caption{Status average 10 fold CV 90/10 ratio}
    \end{subtable}
    
    \begin{subtable}[]{0.5\textwidth}
        \begin{tabular}{ccccc}
        \toprule
        \textbf{CLS} & \textbf{F} & \textbf{P} & \textbf{R} & \textbf{\shortstack{Support Test\\ (10\% of total)}} \\\midrule
        Yes          & 0.931      & 0.935      & 0.926      & 68                                    \\
        No           & 0.872      & 0.889      & 0.880      & 39                                    \\
        Macro        & 0.904      & 0.908      & 0.905      & 109                                   \\
        Weighted     & 0.913      & 0.912      & 0.913      & 109                                  \\\bottomrule
        \end{tabular}
        \caption{Diagnosis average 10 fold CV 90/10 ratio}
    \end{subtable}
\end{table}

\subsection{Meta Annotation Results Breakdown for UCLH}
Aggregate results for each defined meta-annotation at UCLH. Performance is aggregated over all extracted concepts listed in \ref{tab:appdx_concept_table}. We defined the following meta-annotation tasks: 
\begin{itemize}
    \item Negation: is the concept negated or not, values: [Yes / No]
    \item Experiencer: is the concept experienced by the patient or not, values: [Patient, Other]
    \item Problem Temporality: is the concept referring to a historical mention, values [Past Medical Issue, Current Problem]
    \item Certainty: is the concept confirmed to be present, values: [Confirmed, Suspected]
    \item Irrelevant: is the concept relevant, values: [Yes, No]
\end{itemize}

\begin{table}[]
    \caption{Meta Annotation Results at UCLH}
    \begin{subtable}[]{0.5\textwidth}
        \begin{tabular}{ccccc}
        \toprule
        \textbf{CLS} & \textbf{F} & \textbf{P} & \textbf{R} & \textbf{\shortstack{Support Test\\ (10\% of total)}} \\\midrule
        Yes          & 0.896      & 0.895      & 0.900      & 46                                    \\
        No           & 0.688      & 0.767      & 0.631      & 394                                   \\
        Macro        & 0.836      & 0.767      & 0.631      & 440                                   \\
        Weighted     & 0.970      & 0.969      & 0.971      & 440                                  \\\bottomrule
        \end{tabular}
        \caption{Negation: average 10 fold CV 90/10 ratio}
    \end{subtable}
    
    \begin{subtable}[]{0.5\textwidth}
        \begin{tabular}{ccccc}\toprule
        \textbf{CLS} & \textbf{F} & \textbf{P} & \textbf{R} & \textbf{\shortstack{Support Test\\ (10\% of total)}} \\\midrule
        Other        & 0.681      & 0.883      & 0.65       & 3                                     \\
        Patient      & 0.998      & 0.997      & 0.999      & 437                                   \\
        Macro        & 0.940      & 0.940      & 0.825      & 440                                   \\
        Weighted     & 0.996      & 0.996      & 0.996      & 440                                  \\\bottomrule
        \end{tabular}
        \caption{Experiencer: average 10 fold CV 90/10 ratio}
    \end{subtable}
    
    \setlength\tabcolsep{3pt}
    \begin{subtable}[]{0.5\textwidth}
        \begin{tabular}{ccccc}\toprule
        \textbf{CLS}       & \textbf{F} & \textbf{P} & \textbf{R} & \textbf{\shortstack{Support Test\\(10\% of total)}} \\\midrule
        Past Medical Issue & 0.710      & 0.758      & 0.676      & 23                                    \\
        Current Problem    & 0.985      & 0.981      & 0.988      & 412                                   \\
        Macro              & 0.848      & 0.870      & 0.832      & 435                                   \\
        Weighted           & 0.970      & 0.969      & 0.971      & 435                                  \\\bottomrule
        \end{tabular}
        \caption{Problem Temporality: average 10 fold CV 90/10 ratio}
    \end{subtable}
    
    \setlength\tabcolsep{5pt}
    \begin{subtable}[]{0.5\textwidth}
        \begin{tabular}{ccccc}\toprule
        \textbf{CLS} & \textbf{F} & \textbf{P} & \textbf{R} & \textbf{\shortstack{Support Test\\(10\% of total)}} \\\midrule
        Confirmed    & 0.985      & 0.980      & 0.989      & 395                                   \\
        Suspected    & 0.688      & 0.767      & 0.631      & 21                                    \\
        Macro        & 0.836      & 0.874      & 0.810      & 416                                   \\
        Weighted     & 0.970      & 0.970      & 0.971      & 416                                  \\\bottomrule
        \end{tabular}
        \caption{Certainty: average 10 fold CV 90/10 ratio}
    \end{subtable}
    
    \begin{subtable}[]{0.5\textwidth}
        \begin{tabular}{ccccc}\toprule
        \textbf{CLS} & \textbf{F} & \textbf{P} & \textbf{R} & \textbf{\shortstack{Support Test\\(10\% of total)}} \\\midrule
        Yes          & 0.685      & 0.846      & 0.579      & 24                                    \\
        No           & 0.986      & 0.976      & 0.994      & 415                                   \\
        Macro        & 0.835      & 0.911      & 0.787      & 439                                   \\
        Weighted     & 0.969      & 0.970      & 0.972      & 439                                 \\\bottomrule 
        \end{tabular}
        \caption{Irrelevant: average 10 fold CV 90/10 ratio}
    \end{subtable}

\end{table}

\bibliographystyle{elsarticle-num} 

\bibliography{refs}

\end{document}